\documentclass{article}

\usepackage[preprint]{neurips_2025}

\usepackage[utf8]{inputenc} 
\usepackage[T1]{fontenc}    
\usepackage{hyperref}       
\usepackage{url}            
\usepackage{booktabs}       
\usepackage{amsfonts}       
\usepackage{nicefrac}       
\usepackage{microtype}      
\usepackage{xcolor}         

\usepackage{wrapfig,lipsum}
\usepackage{caption}
\usepackage{natbib}
\usepackage{url}
\usepackage{graphicx}
\usepackage{booktabs}
\usepackage{enumitem}
\usepackage{subcaption}
\usepackage{amsmath,amssymb,amsfonts}

\usepackage{xcolor}
\usepackage{mathtools}
\usepackage{dsfont}
\usepackage{hyperref}
\usepackage{bm}
\usepackage{nicefrac}
\usepackage{wrapfig}
\usepackage{lipsum}

\newcounter{assumption}%
\renewcommand{\theassumption}{\arabic{assumption}}

\def\vx{{\bf x}}

\newcommand*\lrp[1]{\left(#1\right)}
\newcommand*\lrn[1]{\left\|#1\right\|}

\def\vx{{\bf x}}
\newcommand{\real}{\ensuremath{\mathbb{R}}}

\usepackage{multirow}
\usepackage{caption}
\usepackage{subcaption}

\usepackage{algorithm, algcompatible}
\usepackage{algpseudocode}
\algnewcommand\algorithmicinput{\textbf{Input:}}
\algnewcommand\INPUT{\item[\algorithmicinput]}
\algnewcommand\algorithmicoutput{\textbf{Output:}}
\algnewcommand\OUTPUT{\item[\algorithmicoutput]}
\algnewcommand\algorithmicoptional{\textbf{Optional:}}
\algnewcommand\OPTIONAL{\item[\algorithmicoptional]}

\usepackage{amsmath,amsfonts,bm}

\def\eqref#1{equation~\ref{#1}}

\def\1{\bm{1}}

\def\vx{{\bm{x}}}

\def\mW{{\bm{W}}}

\DeclareMathAlphabet{\mathsfit}{\encodingdefault}{\sfdefault}{m}{sl}
\SetMathAlphabet{\mathsfit}{bold}{\encodingdefault}{\sfdefault}{bx}{n}

\newcommand{\R}{\mathbb{R}}

\title{Mesh-free sparse identification of nonlinear dynamics}

\author{%
  Mars Liyao Gao \\
  Paul G. Allen School of Computer Science \& Engineering\\
  University of Washington\\
  \texttt{marsgao@uw.edu} \\
  \And
  J. Nathan Kutz \\
  Applied Mathematics and Electrical and Computer Engineering\\
  University of Washington\\
  \texttt{kutz@uw.edu} \\
  \AND
  Bernat Font \\
  Faculty of Mechanical Engineering \\
  Delft University of Technology \\
  \texttt{b.font@tudelft.nl} \\
}

\begin{document}

\maketitle

\begin{abstract}
 Identifying the governing equations of a dynamical system is one of the most important tasks for scientific modeling.
 However, this procedure often requires high-quality spatio-temporal data uniformly sampled on structured grids.
 In this paper, we propose mesh-free SINDy, a novel algorithm which leverages the power of neural network approximation as well as auto-differentiation to identify governing equations from arbitrary sensor placements and non-uniform temporal data sampling.
 We show that mesh-free SINDy is robust to high noise levels and limited data while remaining computationally efficient.
 In our implementation, the training procedure is straight-forward and nearly free of hyperparameter tuning, making mesh-free SINDy widely applicable to many scientific and engineering problems.
 In the experiments, we demonstrate its effectiveness on a series of PDEs including the Burgers' equation, the heat equation, the Korteweg-De Vries equation and the 2D advection-diffusion system.
 We conduct detailed numerical experiments on all datasets, varying the noise levels and number of samples, and we also compare our approach to previous state-of-the-art methods.
 It is noteworthy that, even in high-noise and low-data scenarios, mesh-free SINDy demonstrates robust PDE discovery, achieving successful identification with up to 75\% noise for the Burgers' equation using 5,000 samples and with as few as 100 samples and 1\% noise.
 All of this is achieved within a training time of under one minute.
\end{abstract}

\section{Introduction}
Partial differential equations (PDEs) are one of the most fundamental building blocks of science and engineering.
Through a descriptive symbolic formulation, PDEs can accurately predict spatio-temporal dynamics of different phenomena such as climate, fluid flow, optimal control, or population dynamics, among others.
A-priori knowledge of the system dynamics is required to numerically solve the governing differential equations and obtain an accurate prediction of the system's future state.
However, the explicit form of the governing dynamics can be unknown, so identifying the dynamics of the system is a critical first step for scientific data modeling and prediction~\cite{brunton2016discovering,rudy2017data}.
In this context, the sparse identification of nonlinear dynamics (SINDy) method~\cite{brunton2016discovering} has shown great success in a wide range of applications, and has been extended to improve noise robustness~\cite{Fasel2022} and to provide uncertainty quantification~\cite{Gao2024}.
Still, its applicability has remained limited to space-time structured data since traditional numerical discretization schemes, such as finite differences, are used to compute the candidate \textit{library} terms for the sparse regression on the system dynamics.

\begin{figure}[t]
    \centering
    \includegraphics[width=\linewidth]{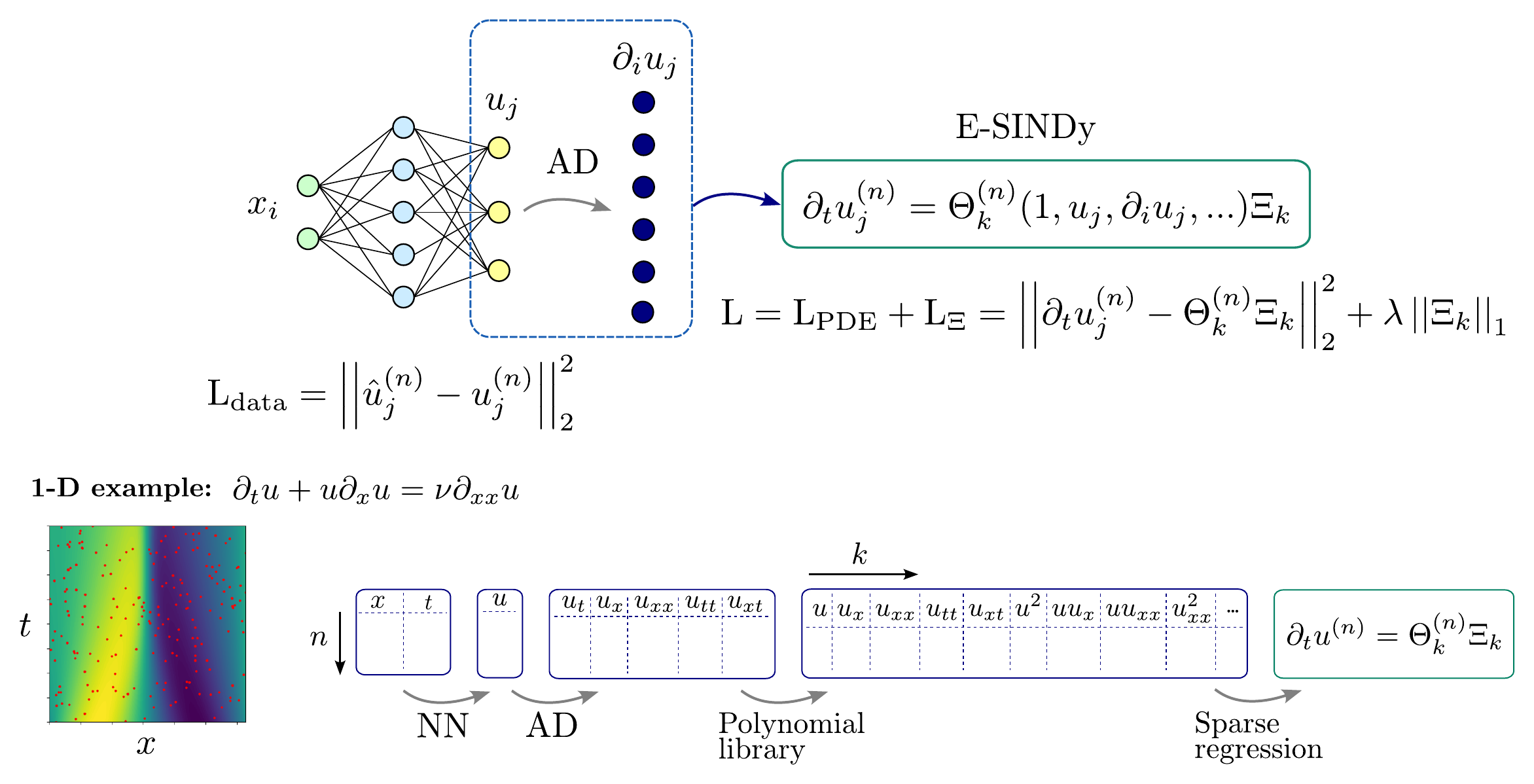}
    \caption{Illustration of the architecture of mesh-free SINDy. From $N$ sensor measurements at random space-time points $x_i^{(n)}$, the neural network approximates the system state $u_j^{(n)}(x_i^{(n)})$. As shown in the 1-D example, the auto-differentiation framework produces $\partial_t u$ (also noted $u_t$), and a library of $K$ candidate terms composed of spatial or mixed derivatives which will be used in the sparse regression.}
    \label{fig:fig1}
\end{figure}

In this paper, we present the mesh-free SINDy algorithm, a novel approach to identify the governing PDE from arbitrary sensor placements.
As illustrated in Fig.~\ref{fig:fig1}, mesh-free SINDy is based on neural network (NN) approximation and auto-differentiation (AD) to  discover the system dynamics from random sensor placement (ie. unstructured data).
The mesh-free SINDy method first uses a NN as a surrogate model of a given unstructured dataset, ie. $f_\mathrm{NN}:x_i\rightarrow u_j$, where $x_i$ is the vector of independent variables and $u_j$ is the NN-approximated system state.
Once the NN is trained according to a loss function $\mathrm{L}(u_j^{(n)},\hat{u}_j^{(n)})$ on $N$ randomly sampled collocation points, a candidates library of $K$ potential terms, $\Theta_k^{(n)}$, is generated via AD by producing point-wise estimates of spatial, temporal, and mixed derivates that may be part of the system dynamics.
Then, the sparse regression step uses ensemble SINDy (E-SINDy) to find the dominant terms of the system dynamics from the library of candidates.

Unlike previous methods which rely on multi-stage training and a combined loss function, we propose to train the NN decoupled from the sparse regression step.
This separation reduces the computational cost of the algorithm while also being robust to high noise levels and small datasets.
Additionally, we avoid setting the weight of each objective in the loss function thus reducing the number of hyperparameters.
In the experiments, we demonstrate the algorithm's efficacy on linear and non-linear dynamical systems of different complexity, comparing our results with previous methods.
Mesh-free SINDy achieves successful model discovery with drastically reduced training cost and minimal hyperparameter tuning, all without sacrificing any performance.
The contribution of our paper is three-fold:
\begin{itemize}
    \item We introduce an AD-based SINDy framework to discover the underlying system dynamics from scarce and randomly sampled data with uncertainty quantification. A shallow neural network is trained as a response surface for the given dataset, and candidate terms of the dynamical system are computed via automatic differentiation.
    \item The training procedure is unified into a single training loop, enabling at least 10x faster execution compared to the traditional workflow, while requiring minimal hyperparameter tuning. This simplified discovery procedure does not compromise accuracy and can even enhance noise robustness in several cases.
    \item We conduct extensive experimental studies on four fundamental PDEs: the Burgers', heat, Korteweg–De Vries, and advection-diffusion equations. Mesh-free SINDy consistently achieves robust discovery from scarce data, even in low-data and high-noise regimes. Detailed experimental setups are provided in Section 4 for full reproducibility.
\end{itemize}

\section{Related works}
The use of NN and AD to approximate the response of dynamical systems from unstructured data was initially proposed by Both et al. \cite{Both2021} through the \textit{DeepMod} framework. Similarly to the present method, derivative terms are generated via AD from the NN response.
DeepMod uses a multi-objective loss function to optimize the models' weights based on (i) the error from the NN system state prediction, $\mathrm{L}_\mathrm{data}=||u_j^{(n)}-\hat{u}_j^{(n)}||^2_2$, (ii) the error from the Lasso-based reconstructed dynamics, $\mathrm{L}_\mathrm{PDE}=||\partial_t u_j^{(n)}-\Theta^{(n)}_k\Xi_k||^2_2$, where $\Xi_k$ is the coefficients vector for each term in $\Theta_k$, and (iii) the regularization term $\mathrm{L}_\mathrm{\Xi}=\lambda||\Xi_k||_1$ to promote sparsity, yielding $\mathrm{L}=\mathrm{L}_\mathrm{data}+\mathrm{L}_\mathrm{PDE}+\mathrm{L}_\mathrm{\Xi}$.
The overall training process, which optimizes both the NN trainable parameters and the $\Xi_k$ coefficients, is performed twice.
The first training includes the regularization term and, after training converges, the $\Xi_k$ vector is reduced to a smaller number of terms according to an arbitrary threshold.
The second training removes the regularization term and is aimed at fine-tunning the remaining $\Xi_k$ coefficients.

Chen et al. \cite{Chen2021} proposed a similar approach to DeepMod, namely \textit{PINN-SR}, which combines both the data loss and the sparse regression of the $\Xi_k$ coefficients in a single loss function.
In that work, authors used an alternating direction optimization (ADO) algorithm instead of the double-training strategy implemented in DeepMod, and the Ridge algorithm instead of Lasso for finding the optimal $\Xi_k$ coefficients.
Recently, Stephany and Earils \cite{Stephany2024} proposed \textit{PDE-LEARN}, a similar framework which also uses a combined loss function of the form $\mathrm{L}=\mathrm{L}_\mathrm{data}+\mathrm{L}_\mathrm{PDE}+\mathrm{L}_\mathrm{\Xi}$.
The novelty in PDE-LEARN relies on the use of rational NNs instead of classic multilayer perceptron architecture, and a triple-stage training strategy consisting of (i) burn-in stage to optimize and prune candidates based on data and PDE losses, (ii) sparsification stage which includes regularization term, and (iii) fine-tuning stage to optimize the coefficients for the surviving terms.

\section{Mesh-free sparse identification of nonlinear dynamics}
A neural network is a nonlinear modeling technique to approximate arbitrary functions.
Given a dataset $\{(x^{(n)}, t^{(n)}, u^{(n)})\}_{n=1}^N$ where $x^{(n)}$ is the spatial location, $t^{(n)}$ is the time, and $u^{(n)}$ is the state variable of the PDE at $(x^{(n)}, t^{(n)})$, the $L$-layer of a neural network can be defined as
\begin{align}
    \hat{u}(x, t):=\sigma_L\lrp{\mW_k\sigma_{k-1}\lrp{\mW_{k-1}\cdots\sigma_1\lrp{\mW_1 \vx_{t-r:t}}}},
\end{align}
where $\theta=\lrp{\mW_1, \dots, \mW_L}$ with $W_1\in\real^{d\times p}$, $\mW_i\in\real^{p\times p}$, $\mW_L\in\real^{p\times 1}$, and $\sigma_i(x)$ represents the activation function.

\paragraph{Mesh-free PDE data} PDEs utilize first principles to model physical and natural phenomena. Given spatial location $x$ and temporal information $t$, the state variable $u$ is a function of the independent spatial and temporal variables, i.e. $u(x, t)$.
Typically, high-quality data collection under fixed spatial grids with regular temporal sampling is required to obtain partial derivative estimates from (e.g.) finite difference methods, which requires $x \in\mathbb{R}^d$ and $t \in\mathbb{R}$ to be structured grids (e.g. $x=[-0.5, -0.4, -0.3, ..., 0.4, 0.5]$ and $t=[0, 0.1, 0.2, ..., 1.0]$).

However, real-world data collection often results in non-uniformly sampled data points in space and time. For instance, uniformly spaced sensors for sea surface temperature data collection are not possible~\citep{reynolds2002improved}.
Hence, a non-uniform dataset of $N$ samples can be defined as $\{(x^{(n)}, t^{(n)}, u^{(n)})\}_{n=1}^N$, where samples are randomly collected within a given spatio-temporal domain.

\paragraph{Automatic differentiation} The automatic differentiation framework is a powerful tool in modern deep learning.
Suppose we model the neural network as $f_\theta(x, t)$, where $\theta$ are the trainable parameters of the neural network
We can compute the spatial and temporal derivatives of the neural network with respect to the input $(x, t)$ using the back-propagation algorithm~\citep{rumelhart1985learning}.
From the computational graph of the back-propagation algorithm, we can estimate spatial, temporal, and mixed derivatives such as $\frac{\partial u}{\partial t}, \frac{\partial u}{\partial x}, \frac{\partial^2 u}{\partial x\partial t}, \frac{\partial^2 u}{\partial x^2}, \frac{\partial^3 u}{\partial x^3}$, and so on.

\paragraph{Partial derivative estimation from mesh-free data} Compared to traditional numerical differentiation methods, AD can obtain direct space-time derivatives of a non-uniform dataset from a fitted analytical response surface, such as a neural network.
This mitigates interpolation errors that otherwise arise when projecting non-uniformly sampled data on structured grids.
Adapting from the better estimated derivatives, we then perform sparse regression to discover the governing PDE via SINDy and its variants~\citep{brunton2016discovering,Fasel2022,bertsimas2023learning,gao2023convergence}.

We can now perform PDE discovery with arbitrary sensor placements with irregularly-sampled time series~\citep{rubanova2019latent}, which can relax the complexity of data collection and reduce the cost of data acquisition~\citep{iyer2019living}, as presented in Alg.~\ref{alg:mesh_free_sindy}.

\begin{algorithm*}[t]
    \caption{\textsc{Mesh–free SINDy} (sparse identification of nonlinear dynamics)}
    \label{alg:mesh_free_sindy}
    \begin{algorithmic}[1]
        \Require data $\mathcal{D}=\{(x^{(n)},t^{(n)},u^{(n)})\}_{n=1}^{N}$,
                dictionary map $\Theta$, sparsity threshold $\lambda$
        \Ensure  sparse coefficient matrix $\Xi$
        \smallskip
        \State initalize a neural surrogate $f_\theta:\R^{d}\!\times\!\R\to\R$  \Comment{random weights}
        \Repeat
            \State $\displaystyle
                \theta \;\leftarrow\;
                \arg\min_\theta
                    \frac1N\sum_{n=1}^{N}\lrn{u^{(n)}-f_\theta(x^{(n)},t^{(n)})}^{2}$ \Comment{MSE training}
        \Until{convergence}
        \smallskip
        \For{$n=1,\dots,N$}
            \State $\hat u^{(n)}          \;\leftarrow\; f_\theta(x^{(n)},t^{(n)})$                \Comment{forward pass}
            \State $\partial_t\hat u^{(n)} \leftarrow\; \partial_t f_\theta(x^{(n)},t^{(n)})$     \Comment{auto-diff}
        \EndFor
        \smallskip
        \State $\mathbf{\Theta} \;\leftarrow\;
               \left[
                 \Theta\lrp{\hat u^{(n)},\,\partial_x\hat u^{(n)},\,\partial_x^{2}\hat u^{(n)},\dots}
               \right]_{n=1}^{N}$                                              \Comment{build dictionary}
        \smallskip
        \State $\displaystyle
               \Xi \;\leftarrow\;
               \arg\min_{\Xi}
               \frac1N\,
               \bigl\|\partial_t\hat u - \mathbf{\Theta}\Xi\bigr\|_{2}^{2}
               +\lambda\,\|\Xi\|_{1}$                                         \Comment{sparse regression}
        \State \Return $\Xi$
    \end{algorithmic}
\end{algorithm*}

\paragraph{Qualifying PDE discovery uncertainty from automatic differentiation}
Uncertainty quantification is an essential step in scientific machine learning and model discovery as it provides confidence of discovery as well as safe predictive inference for the discovered model.
This is particularly important for critical applications such as health-care, climate modeling, or aerodynamic design, among others.

To enable uncertainty estimate for mesh-free SINDy,
we perform non-parametric bootstrap via the following procedure.
Firstly, we randomly subsample $m<N$ data points from the original dataset $\mathcal{D}=\{\mathbf{\Theta}, \mathbf{\hat{u}}_t\}$ without replacement, denoted as $\Tilde{\mathcal{D}}$, where $\mathbf{\Theta}=\Theta\lrp{\hat{u}^{(n)},\,\partial_x\hat{u}^{(n)},\,\partial_x^{2}\hat{u}^{(n)},\dots}$ and $\mathbf{\hat{u}}_t=\partial_t \hat{u}_\theta(x^{(n)},t^{(n)})$.
Then, the algorithm runs SINDy on the bootstrapped dataset $\Tilde{\mathcal{D}}$ to perform sparse regression and obtain the PDE coefficients $\Tilde{\Xi}$.
Finally, repeating the above procedure $M$ times to obtain $M$ bootstrapped PDE coefficients $\{\Tilde{\Xi}^m\}_{m=1}^M$.

The Bootstrapping algorithm above can obtain an empirical distribution $\{\Tilde{\Xi}^m\}_{m=1}^M$, which enables uncertainty quantification for mesh-free SINDy.
This algorithm can be implemented via ensemble SINDy~\citep{Fasel2022}, which is theoretically grounded to provide asymptotically correct distributional estimates~\citep{Gao2024}.
In the experiments, we provide the uncertainty quantification results for the discovered PDEs via mesh-free SINDy.

\section{Numerical experiments}
In the following sections, we present case studies across a range of fundamental PDEs, including the Burgers' equation, the heat equation, the Korteweg–De Vries equation, and the 2D advection-diffusion equation. We summarize and highlight the experimental results in Tab.~\ref{tab:pde_summary}. Further details about each experiment are provided in the subsequent subsections.

\begin{table}[htbp]
    \centering
    \scriptsize
    \caption{Performance of Mesh-free SINDy on tested PDEs}
    \label{tab:pde_summary}
    \resizebox{\columnwidth}{!}{
    \begin{tabular}{cccccc}
        \toprule
        \textbf{PDE} & \textbf{Sample} & \textbf{Noise} & \textbf{Success} & \textbf{Time (s)} & \textbf{Recovered Equation} \\
        \midrule
        \multirow{8}{*}{\begin{tabular}{c}
            \textbf{Burgers'}\\
            \includegraphics[width=0.12\textwidth]{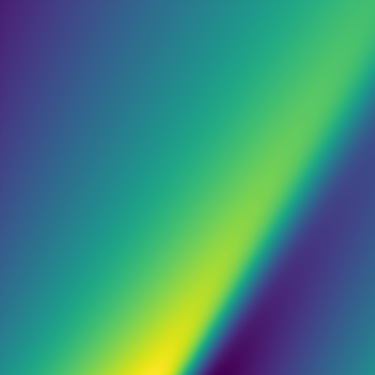}\\
            $u_t + uu_x = 0.5 u_{xx}$
        \end{tabular}}
         & 100 & 1\% & 66.7\% & 18 & $u_t + 1.03uu_x = 0.59 u_{xx}$ \\
            & 1000 & 1\% & 100.0\% & 20 & $u_t + 0.98uu_x = 0.47 u_{xx}$ \\
            & 2500 & 1\% & 100.0\% & 20 & $u_t + 1.00uu_x = 0.49 u_{xx}$ \\
            & 5000 & 1\% & 100.0\% & 21 & $u_t + 1.00uu_x = 0.49 u_{xx}$ \\
            & 100 & 10\% & 50.0\% & 25 & $u_t + 0.98uu_x = 0.41 u_{xx}$ \\
            & 1000 & 10\% & 100.0\% & 25 & $u_t + 0.94uu_x = 0.39 u_{xx}$ \\
            & 2500 & 50\% & 80.0\% & 25 & $u_t + 0.79uu_x = 0.37 u_{xx}$ \\
            & 5000 & 75\% & 80.0\% & 25 & $u_t + 0.73uu_x = 0.37 u_{xx}$ \\
            \midrule
            \multirow{8}{*}{\begin{tabular}{c}
                \textbf{Heat}\\
                \includegraphics[width=0.12\textwidth]{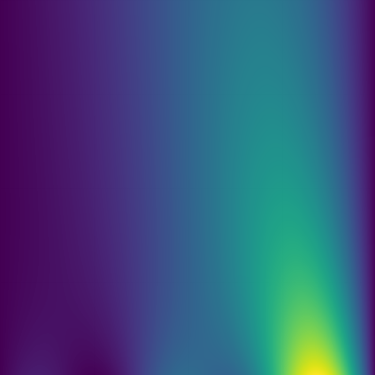}\\
                $u_t = u_{xx}$
            \end{tabular}}
            & 500 & 1\% & 8.3\% & 29 & $u_t = 0.35 u_{xx}$ \\
            & 1000 & 1\% & 91.7\% & 29 & $u_t = 0.59 u_{xx}$ \\
            & 2500 & 1\% & 91.7\% & 30 & $u_t = 0.75 u_{xx}$ \\
            & 4000 & 1\% & 100.0\% & 30 & $u_t = 0.85 u_{xx}$ \\
            & 500 & 5\% & 8.3\% & 29 & $u_t = 0.27 u_{xx}$ \\
            & 1000 & 10\% & 91.7\% & 29 & $u_t = 0.48 u_{xx}$ \\
            & 2500 & 25\% & 8.3\% & 30 & $u_t = 0.43 u_{xx}$ \\
            & 4000 & 25\% & 50.0\% & 30 & $u_t = 0.33 u_{xx}$ \\
            \midrule
            \multirow{8}{*}{\begin{tabular}{c}
                \textbf{KdV}\\
                \includegraphics[width=0.12\textwidth]{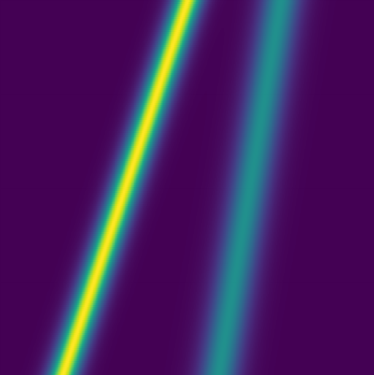}\\
                $u_t + 6uu_x + u_{xxx} = 0 $
            \end{tabular}}
            & 100 & 5\% & 8.3\% & 48 & $u_t + 1.2uu_x + 0.18u_{xxx} = 0$ \\
            & 500 & 5\% & 25.0\% & 52 & $u_t + 5.75uu_x + 0.98u_{xxx} = 0$ \\
            & 1000 & 5\% & 75.0\% & 56 & $u_t + 5.38uu_x + 0.84u_{xxx} = 0$ \\
            & 2000 & 5\% & 83.3\% & 69 & $u_t + 5.89uu_x + 1.00u_{xxx} = 0$ \\
            & 100 & 10\% & 0.0\% & 48 & $u_t + 2.34uu_x + 1.06u_{x} = 0$ \\
            & 500 & 10\% & 50.0\% & 52 & $u_t + 5.31uu_x + 0.79u_{xxx} = 0$ \\
            & 1000 & 25\% & 41.6\% & 56 & $u_t + 5.03uu_x + 0.73u_{xxx} = 0$ \\
            & 2000 & 50\% & 66.7\% & 40 & $u_t + 5.02uu_x + 0.84u_{xxx} = 0$ \\
            \midrule
            \multirow{8}{*}{\begin{tabular}{c}
                \textbf{Advection-Diffusion}\\
                \includegraphics[width=0.12\textwidth]{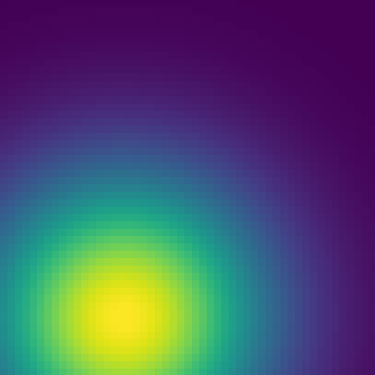}\\
                $\partial_t u = -\nabla \cdot \left( \vec{c} u \right) + 0.5 \nabla^2 u$
            \end{tabular}}
            & 500 & 10\% & 100.0\% & 31 & $u_t + 0.24 u_x + 0.49 u_y = 0.49u_{xx} + 0.49u_{yy}$ \\
            & 2000 & 10\% & 100.0\% & 33 & $u_t + 0.25 u_x + 0.50 u_y =  0.49u_{xx} + 0.49u_{yy}$ \\
            & 4000 & 10\% & 100.0\% & 36 & $u_t + 0.25 u_x + 0.51 u_y =  0.48u_{xx} + 0.49u_{yy}$ \\
            & 6000 & 10\% & 100.0\% & 38 & $u_t + 0.25u_x + 0.50u_y =  0.50u_{xx} + 0.48u_{yy}$ \\
            & 500 & 25\% & 8.3\% & 31 & $u_t + 0.27 u_x + 0.50 u_y =  0.45u_{xx} + 0.48u_{yy}$ \\
            & 2000 & 25\% & 100.0\% & 33 & $u_t + 0.24 u_x + 0.49 u_y =  0.47u_{xx} + 0.47u_{yy}$ \\
            & 4000 & 50\% & 91.6\% & 36 & $u_t + 0.24 u_x + 0.51 u_y =  0.45u_{xx} + 0.48u_{yy}$ \\
            & 6000 & 75\% & 25.0\% & 38 & $u_t + 0.28 u_x + 0.48 u_y =  0.49u_{xx} + 0.45u_{yy}$\\
            \bottomrule
            \end{tabular}
    }
\end{table}

\subsection{Burgers' equation}
The Burgers' equation is a fundamental nonlinear PDE that can describe turbulence and viscous fluid flow~\cite{bateman1915some,burgers1948mathematical} in 1D with the combination of nonlinear advection and diffusion.

The governing equation is given by
\begin{align}
    u_t + uu_x = \nu u_{xx},
\end{align}
where $u$ is the velocity, $\nu$ is the viscosity, $u_t$ is the temporal derivative, and $u_x, u_{xx}$ are first and second order spatial derivatives, respectively.

In the experiments, we set the viscosity $\nu=0.5$ with a periodic boundary conditions for the domain $x\in[0, 2\pi]$ and $t\in[0, 1]$ where
\begin{align}
    u(0, t) = u(2\pi, t).
\end{align}
We simulate the PDE with random sensor placement $\{(x^{(n)}, t^{(n)})\}_{n=1}^N$ where $N$ is the sample size.
To further simulate the PDE, we follow the closed-form solution of the viscous Burgers' equation
\begin{align}
    u(x, t) = 4 - 2\nu \frac{\frac{d}{dx}\phi(x, t)}{\phi(x, 0)},
\end{align}
where $\phi(x,t) = \exp\!\left(-\frac{(x-4t)^2}{\left(4\nu(t+1)\right)^2}\right)
+ \exp\!\left(-\frac{(x-4t-2\pi)^2}{\left(4\nu(t+1)\right)^2}\right)$.

We setup a fully-connected two-layer neural network with 32 neurons per layer.
For continuous differentiation, we utilize the $\mathrm{tanh}$ activation function.
The neural network is trained with the AdamW optimizer with a learning rate of $0.0005$, weight decay of $0.01$, and a batch size of $20$.
We set the training epoch to $200$ which approximately takes 25 seconds to run without GPU acceleration.
To identify the system dynamics with SINDy, we set the interaction order to $2$, which includes 20 polynomial terms conformed of
$u, u_x, u_{xx}, u_{tt}, u_{xt}, u^2$ and their interactions.
We utilize an ensemble sequentially thresholding least-squares algorithm with threshold 0.14 and $\ell_2$ regularization 0.05.
These hyperparameters are kept constant during the experiments when varying dataset size and noise level.

The training convergence as well as the neural network response surface are included in Appendix~\ref{app:burgers}.
We show in Tab.~\ref{tab:burgers_success} that the mesh-free SINDy algorithm achieves robust discovery with up to 50\% noise using 5,000 samples, with a 100\% success rate.
With 75\% injected noise, we achieve an 80\% success rate with 5,000 samples.
In the low-data limit, we achieve a 66.7\% success rate with 100 samples and 1\% noise, and a 50\% success rate with 10\% noise, which is notably robust when testing non-uniformly sampled data.

\begin{figure}
    \centering
    \includegraphics[width=\linewidth]{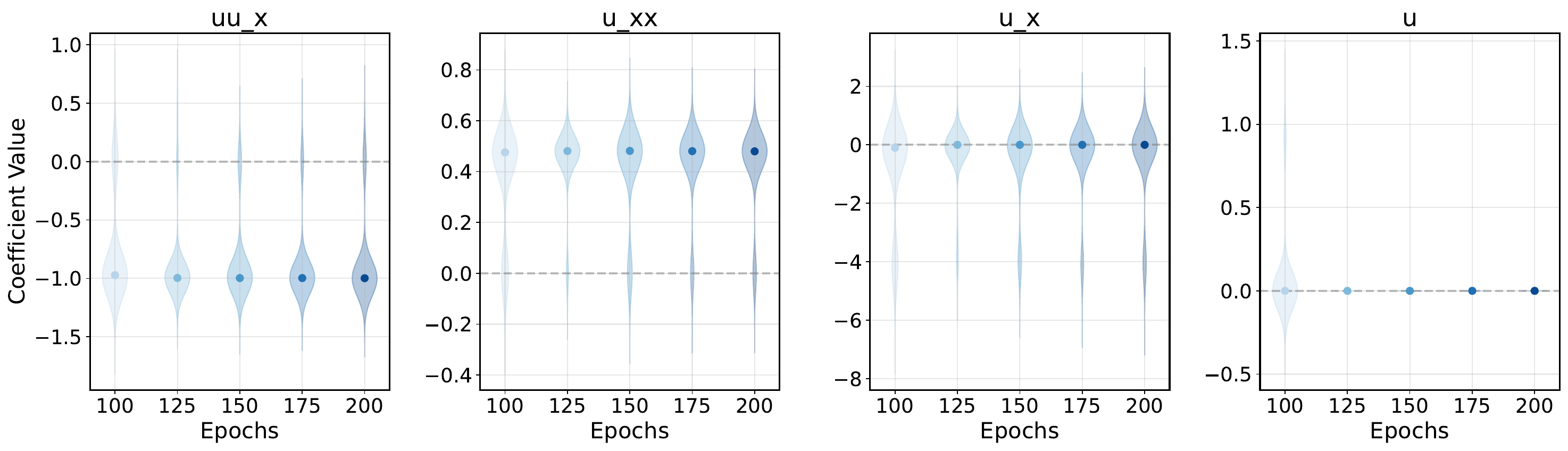}
    \caption{Uncertainty quantification and epoch-wise evaluation of the discovered SINDy model for the Burgers' equation.}
    \label{fig:burgers_sindy_evolution}
\end{figure}

Fig.~\ref{fig:burgers_sindy_evolution} presents the uncertainty estimate of mesh-free SINDy, where ensemble SINDy discovery is performed every 25 epochs.
This figure demonstrates the distribution estimation of mesh-free SINDy with an increasing number of training epochs.
It is worth noting that insufficient initial training can lead to incorrect model discovery, but the model will converge to the correct physics with more training epochs. In turn, it demonstrates that longer training periods leads to more accurate derivative estimation in results in a higher success rate of PDE discovery.

\begin{table}[t]
    \centering
    \caption{PDE discovery success rate (\%) under different sample sizes and noise levels for the Burgers' equation.}
    \label{tab:burgers_success}
    \begin{tabular}{cccccc}
        \toprule
        \textbf{Samples} & \textbf{1\% Noise} & \textbf{10\% Noise} & \textbf{25\% Noise} & \textbf{50\% Noise} & \textbf{75\% Noise} \\
        \midrule
        100   & 66.7 & 50.0 & 16.7 & 16.7 & 0.0 \\
        1000  & 100.0 & 100.0 & 50.0 & 25.0 & 0.0 \\
        2500  & 100.0 & 100.0 & 100.0 & 80.0 & 40.0 \\
        5000  & 100.0 & 100.0 & 100.0 & 100.0 & 80.0 \\
        \bottomrule
    \end{tabular}
\end{table}

\subsection{1D heat equation}
The heat equation is a fundamental PDE that characterizes the heat transfer via a diffusion process.
In 1D, the heat equation can be written as
\begin{align}
    u_t = \nu u_{xx},
\end{align}
where $u$ is the temperature and $\nu=1.0$ is the diffusion coefficient. We set up the initial condition to be $u(x,0)=x^2\sin(x)\approx \sum_q D_q\sin\lrp{q x}$ in a $x\in[0,\pi]$ domain with periodic boundary conditions, i.e. $u(0, t)=u(\pi,t)$.

In the experiments, we use a neural network with four fully-connected layers of 128 neurons each, again using $\mathrm{tanh}$ as the activation function.
A batch size of 20 samples is considered, and the learning rate is set to $2e^{-4}$ for $300$ training epochs.
The MSE loss function between target and predicted data is applied.
Following the previous example, we set the interaction order to $3$ thus including 12 polynomial terms
\[
1, u, u_x, u_{xx}, u_{xxx}, u u, u u_x, uu_{xx}, uu_{xxx}, u^2u_x, u^2u_{xx}, u^2u_{xxx}.
\]
We utilize mixed-integer optimizer to perform sparse regression with $\ell_0$ regularization with $\ell_2=0.05$.
These hyperparameters are kept constant during the experiments when varying dataset size and noise level.

\begin{figure}[t]
    \centering
    \includegraphics[width=\linewidth]{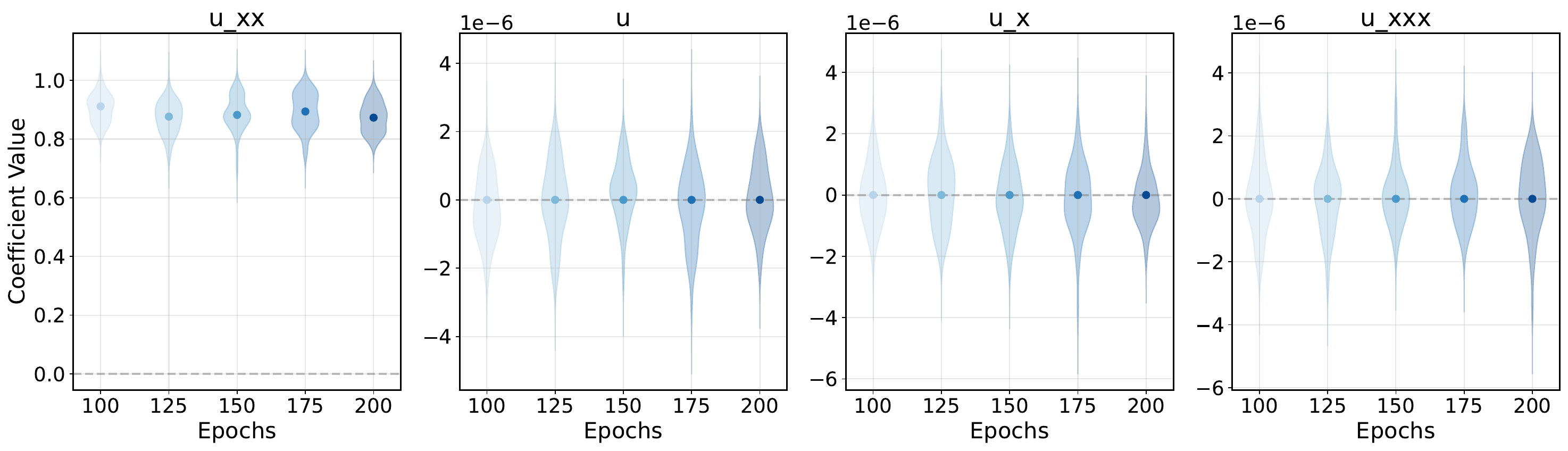}
    \caption{Uncertainty quantification and epoch-wise evaluation of the discovered SINDy model for the heat equation.}
    \label{fig:heat_sindy_evolution}
\end{figure}

The training convergence as well as the neural network response surface are included in Appendix~\ref{app:heat}.
In Tab.~\ref{tab:heat_success}, mesh-free SINDy achieves robust discovery with up to 1\% noise using 4,000 samples, with a 100\% success rate.
With 25\% injected noise, we achieve a 50\% success rate with 4,000 samples.
In the low-data limit, we achieve a 91.7\% success rate with 1000 samples and 1\% noise.
Remarkably, the model can still successfully discover the heat equation with only 100 samples under a noisy environment.
Fig.~\ref{fig:heat_sindy_evolution} presents the uncertainty estimate of mesh-free SINDy, where ensemble SINDy discovery is performed every 25 epochs.
The correct term $u_{xx}$ is robustly discovered with statistical significance, while other terms are concentrated around zero.

\begin{table}[t]
    \centering
    \caption{PDE discovery success rate (\%) under different sample sizes and noise levels for the heat equation.}
    \label{tab:heat_success}
    \begin{tabular}{ccccc}
        \toprule
        \textbf{Samples} & \textbf{1\% Noise} & \textbf{5\% Noise} & \textbf{10\% Noise} & \textbf{25\% Noise}\\
        \midrule
        500   & 8.3 & 8.3 & 8.3 & 0.0\\
        1000  & 91.7 & 83.3 & 91.7 & 0.0  \\
        2500  & 91.7 & 91.7 & 61.7 & 8.3 \\
        4000  & 100.0 & 75.0 & 66.7 & 50.0 \\
        \bottomrule
    \end{tabular}
\end{table}

\subsection{Korteweg–De Vries equation}
The Korteweg–De Vries (KdV) equation is a mathematical model of waves on shallow water surfaces.
The equation is a third-order nonlinear PDE which can be written as
\begin{align}
    u_t + uu_x + u_{xxx} = 0,
\end{align}
where $u$ is the wave height and $u_x$ and $u_t$ are the spatial and temporal derivatives.

In the experiments, we apply the two-soliton solution of the KdV equation with the initial condition being a superposition with functions of the form
\begin{align}
    u(x, t) = \frac{c}{2}\text{sech}^2\lrp{\frac{\sqrt{c}}{2}(x-ct-a)},
\end{align}
with offset centers and different amplitudes following~\citep{rudy2017data}.

In the experiments, we use a fully-connected neural network with four layers and 30 neurons per layer.
The $\mathrm{tanh}$ activation function is employed.
We use the Adam optimizer with learning rate $1e^{-3}$ and apply full-batch gradient descent as the mesh-free dataset is small.
The training time for $500$ samples and $10,000$ epochs is around 50 seconds using the MSE loss.
The training convergence as well as the neural network response surface are included in Appendix~\ref{app:kdv}.
Following the previous example, we set the following library for discovery:
$
1, u, u_x, u_{xx}, u_{xxx}, u u, u u_x, uu_{xx}, uu_{xxx}, u^2u_x, u^2u_{xx}, u^2u_{xxx}.
$
We utilize mixed-integer optimizer to perform sparse regression with $\ell_0$ regularization with $\ell_2=0.05$.
These hyperparameters are kept constant during the experiments when varying dataset size and noise level.

\begin{figure}[t]
    \centering
    \includegraphics[width=\linewidth]{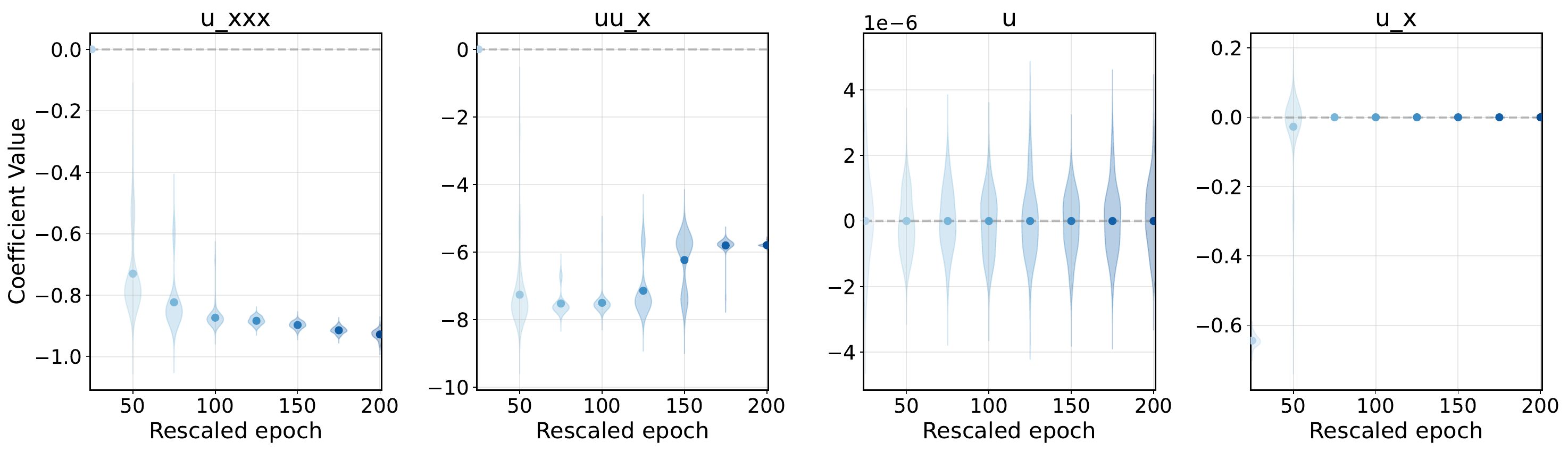}
    \caption{Uncertainty quantification and epoch-wise evaluation of the discovered SINDy model for the KdV equation.}
    \label{fig:kdv_sindy_evolution}
\end{figure}

Mesh-free SINDy achieves robust discovery under 1\% noise using 2,000 samples with a 100\% success rate.
In Tab.~\ref{tab:kdv_success}, with 50\% injected noise, we achieve a 66.7\% success rate with 2,000 samples.
In the low-data limit, we achieve a 50\% success rate with 500 samples and 5\% noise.
Remarkably, the model can still successfully discover the KdV equation with only 100 samples under 5\% noise.
Fig.~\ref{fig:kdv_sindy_evolution} presents the uncertainty estimate of mesh-free SINDy, where ensemble SINDy discovery is performed every 2500 epochs.
The correct terms, $u_{xxx}$ and $uu_x$, are robustly discovered with uncertainty estimates, while other terms consistently evolve into a distribution centered at zero.

\begin{table}[t]
    \centering
    \caption{PDE discovery success rate (\%) under different sample sizes and noise levels for the KdV equation.}
    \label{tab:kdv_success}
    \begin{tabular}{ccccc}
        \toprule
        \textbf{Samples} & \textbf{5\% Noise} & \textbf{10\% Noise} & \textbf{25\% Noise} & \textbf{50\% Noise}\\
        \midrule
        100   & 8.3 & 0.0 & 0.0 & 0.0 \\
        500  & 50.0 & 25.0 & 16.6 & 8.3 \\
        1000  & 75.0 & 50.0 & 41.6 & 33.3 \\
        2000  & 83.3 & 75.0 & 83.3 & 66.7 \\
        \bottomrule
    \end{tabular}
\end{table}

\subsection{2D advection-diffusion equation}
The two-dimensional advection-diffusion equation combines advection and diffusion processes in two spatial dimensions.
Advection-diffusion equations are widely used to describe many linear processes in the physical sciences, including heat transfer, pollutant transport, and fluid dynamics. The equation is given by
\begin{align}
    \partial_t u = -\nabla \cdot \left( \vec{c} u \right) + K \nabla^2 u,
\end{align}
where $u$ is the concentration, $\vec{c}=(0.25, 0.5)$ is the velocity vector, and $K$ is the diffusion coefficient.
We set the initial condition as a Gaussian distribution.
In the experiments, we use a neural network with eight fully-connected layers and 60 neurons per layer.
The $\mathrm{tanh}$ activation function is employed.
We use the Adam optimizer with learning rate $1e^{-3}$ and apply full-batch gradient descent as the mesh-free dataset is small.
The training time for $500$ samples and $10,000$ epochs is around 30 seconds under the MSE loss.
We set the following library for discovery:
$1, u, u^2, u^3, u_x, u_y, u_{xx}, u_{yy}, uu_x, uu_y.$
We utilize mixed-integer optimizer to perform sparse regression with $\ell_0$ regularization with $\ell_2=0.0005$.
These hyperparameters are kept constant during the experiments when varying dataset size and noise level.

\begin{figure}[t]
    \centering
    \includegraphics[width=\linewidth]{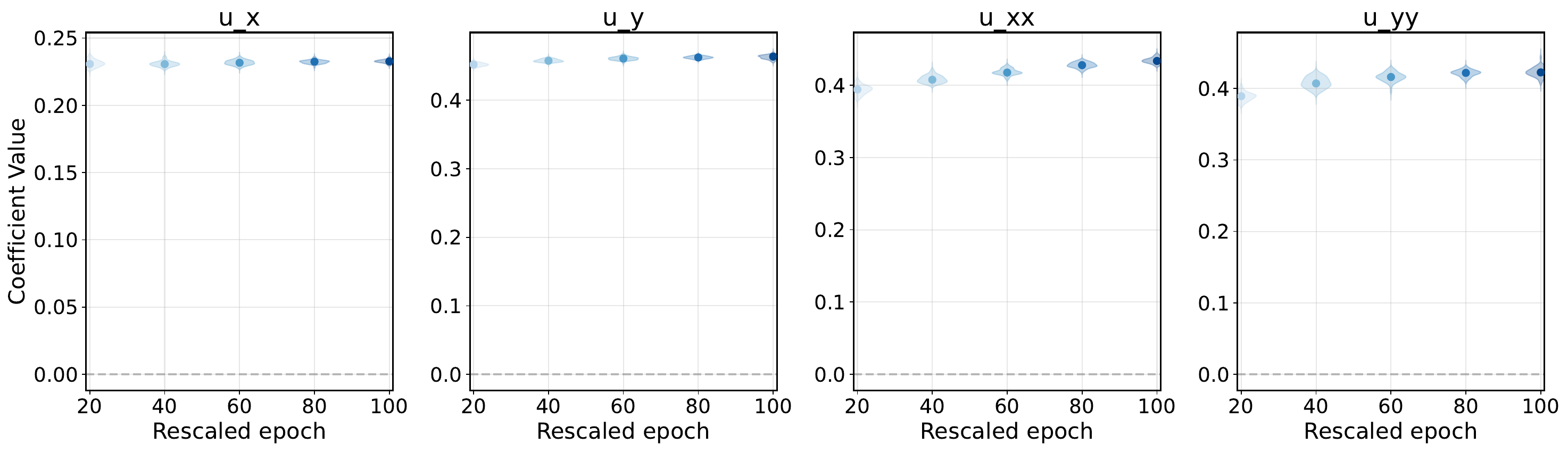}
    \caption{Uncertainty quantification and epoch-wise evaluation of the discovered SINDy model for the 2D advection-difusion equation.}
    \label{fig:ad_sindy_evolution}
\end{figure}

The training convergence as well as the neural network response surface are included inAppendix~\ref{app:ad}.
Mesh-free SINDy achieves robust discovery under 25\% noise using 6,000 samples with a 100\% success rate.
In Tab.~\ref{tab:ad_success}, with 75\% injected noise, we achieve a 25\% success rate with 6,000 samples.
In the low-data limit, we achieve a 100\% success rate with 500 samples and 10\% noise.
Remarkably, the model can still successfully discover the advection-diffusion equation with only 500 samples up to 25\% noise.
Fig.~\ref{fig:ad_sindy_evolution} presents the uncertainty estimate of mesh-free SINDy, where ensemble SINDy discovery is performed every 2500 epochs.
The correct terms are robustly discovered with uncertainty estimates.

\subsection{Comparison with previous works}
\begin{table}[t]
    \centering
    \caption{PDE discovery success rate (\%) under different sample sizes and noise levels for the 2D advection-difussion equation.}
    \label{tab:ad_success}
    \begin{tabular}{cccccc}
        \toprule
        \textbf{Samples} & \textbf{1\% Noise} & \textbf{10\% Noise} & \textbf{25\% Noise} & \textbf{50\% Noise} & \textbf{75\% Noise}\\
        \midrule
        500   & 100.0 & 100.0 & 8.3 & 0.0 & 0.0 \\
        2000  & 100.0 & 100.0 & 100.0 & 16.7 & 0.0 \\
        4000  & 100.0 & 100.0  & 100.0  & 91.6 & 8.3 \\
        6000  & 100.0 & 100.0 & 100.0 & 91.6 & 25.0 \\
        \bottomrule
    \end{tabular}
\end{table}

\begin{table}[t]
    \centering
    \caption{Comparison of mesh-free SINDy with previous methods for the Burgers' equation experiment.}
    \label{tab:comparison}
    \begin{tabular}{lcccccc}
        \toprule
        \textbf{Method} & \textbf{Cost} & \textbf{Noise level} & \textbf{Samples} & \textbf{Tuning} & \textbf{Success rate} & \textbf{Data Collection} \\
        \midrule
        Mesh-free SINDy & 30s & 100\% & 100 & Low & High & Arbitrary \\
        EW-SINDy \citep{Fasel2022} & 1s & 100\% & 25,000 & Low & High & Grid \\
        PINN-SR \citep{Chen2021} & 20m & 10\% & 800 & Medium & Medium & Arbitrary \\
        PDE-LEARN \citep{Stephany2024} & 30m & 100\% & 250 & High & Medium & Arbitrary \\
        DeepMOD \citep{Both2021} & 4m & 75\% & 100 & High & High & Arbitrary \\
        \bottomrule
    \end{tabular}
\end{table}

The performance of mesh-free SINDy is compared with previous methods using the Burgers' equation as a baseline.
Other methods have been run according to the documentation and best-practices available in their open-source repositories.
All methods are tested under the same GPU hardware.

Tab.~\ref{tab:comparison} summarizes the results for each method.
In general, mesh-free SINDy offers the simplest implementation with fast computation and minimal tuning, while maintaining high robustness to noise, especially in low-sample scenarios.
Ensemble Weak (EW-)SINDy~\citep{Fasel2022,messenger2021weak} can tolerate very high noise levels, but it requires grid-based data collection with a large sample size, which is not always feasible in real-world applications.
PINN-SR and PDE-LEARN both have relatively slow training times and require careful tuning of hyperparameters.
This can be particularly challenging in the context of deep learning, which adds extra complexity on top of the neural network tuning required for optimal training.
In additon, reproducibility is also hindered by the fine-tunning procedure, which can be nontrivial and biased towards prior knowledge.
DeepMOD has a fast training time and is robust to noise under very small sample sizes. However, the tuning effort required is high, as one potentially needs to tune the threshold for different levels of noise.
The periodic thresholding scheduler also requires careful manipulation, which can be challenging in practice.

On the other hand, mesh-free SINDy has encountered limitations to discover very high-order PDEs, such as the Kuramoto–-Sivashinsky equation, which contains a 4th-order derivative in space.
Higher order terms may contain errors as a result of recursive (nested) first-order AD \citep{Pearlmutter2007,Karczmarczuk2001}.
Other methods enforce strong priors during the early pruning of the candidate library as part of a multi-stage training approach, which helps focus the sparse regression step on a reduced set of terms.
Hence, future work will be focused on improving the AD step to produce better estimates of high-order derivatives.

\section{Conclusion}
In this paper we present a novel algorithm to discover governing equations from non-uniformly sampled and scarce data, namely mesh-free SINDy.
Based on neural networks and automatic differentiation, mesh-free SINDy demonstrates a remarkable ability to uncover the hidden system dynamics under noisy environments.
In particular, we test the algorithm for the following systems: 1D viscous Burgers' equation, 1D heat equation, Korteweg–De Vries equation, and 2D advection-diffusion equation.
In most cases, the success rate for recovering the system governing equations is consistently high.
For high noise levels (up to 100\%), mesh-free SINDy can still recover the correct dynamics given enough data.

Differently from previous works, the present approach decouples the loss from the data fitting and the PDE residual, and we also use the ensemble SINDy algorithm to provide uncertainty quantification on the discovered dynamics.
This enables mesh-free SINDy to be both robust to noise and small datasets, while also substantially lowering the overall training cost.

\bibliographystyle{abbrv}
\bibliography{references}

\appendix

\section{Experimental details}
\subsection{Error Metrics}
We evaluate our SINDy model using five complementary error metrics:

\begin{itemize} \item \textbf{PDE Conservation Error} ($E_{\text{PDE}}$): Measures how well the discovered model satisfies the underlying partial differential equation conservation laws
\begin{equation}
    E_{\text{PDE}} = \frac{1}{N} \sum_{n=1}^{N} \left| \frac{\partial u^{(n)}}{\partial t} - \mathcal{F}_{\text{SINDy}}(u^{(n)}, \nabla u^{(n)}, \nabla^2 u^{(n)}, \ldots) \right|_2^2
\end{equation}
where $\mathcal{F}_{\text{SINDy}}$ represents the identified sparse PDE model.

\item \textbf{Neural Network Residual} ($E_{\text{NN}}$): Quantifies the fitting accuracy of the neural network to the training data
\begin{equation}
    E_{\text{NN}} = \frac{1}{N} \sum_{n=1}^{N} \left| u^{(n)} - \hat{u}_{\text{NN}}(x^{(n)}, t^{(n)}) \right|_2^2
\end{equation}
where $\hat{u}_{\text{NN}}$ is the neural network approximation of the state.

\item \textbf{Temporal Derivative Error} ($E_{\text{dudt}}$): Evaluates the accuracy in predicting the time derivatives
\begin{equation}
    E_{\text{dudt}} = \frac{1}{N} \sum_{n=1}^{N} \left| \frac{\partial u^{(n)}}{\partial t} - \frac{\partial \hat{u}_{\text{NN}}(x^{(n)}, t^{(n)})}{\partial t} \right|_2^2
\end{equation}
where $\frac{\partial \hat{u}_{\text{NN}}}{\partial t}$ is the time derivative computed from the neural network computed with automatic differentiation.

\item \textbf{SINDy Prediction Error} ($E_{\text{SINDy}}$): Measures how well the sparse identified model approximates the right-hand side dynamics
\begin{equation}
    E_{\text{SINDy}} = \frac{1}{N} \sum_{n=1}^{N} \left| \frac{\partial \hat{u}_{\text{NN}}(x^{(n)}, t^{(n)})}{\partial t} - \mathcal{F}_{\text{SINDy}}(\hat{u}^{(n)}_{\text{NN}}, \nabla \hat{u}^{(n)}_{\text{NN}}, \nabla^2 \hat{u}^{(n)}_{\text{NN}}, \ldots) \right|_2^2
\end{equation}
comparing the neural network time derivatives with the SINDy model predictions.

\item \textbf{State Reconstruction Error} ($E_{\text{field}}$): Quantifies the accuracy in reconstructing the solution field
\begin{equation}
    E_{\text{field}} = \frac{1}{N} \sum_{n=1}^{N} \left| u^{(n)} - \hat{u}^{(n)}_{\text{NN}} \right|_2^2
\end{equation}
where $u^{(n)}$ represents the ground truth solution values, potentially from high-fidelity simulations or experimental data.
\end{itemize}

\subsection{Burgers' equation}
\label{app:burgers}
\begin{figure}[H]
    \centering
    \includegraphics[width=\linewidth]{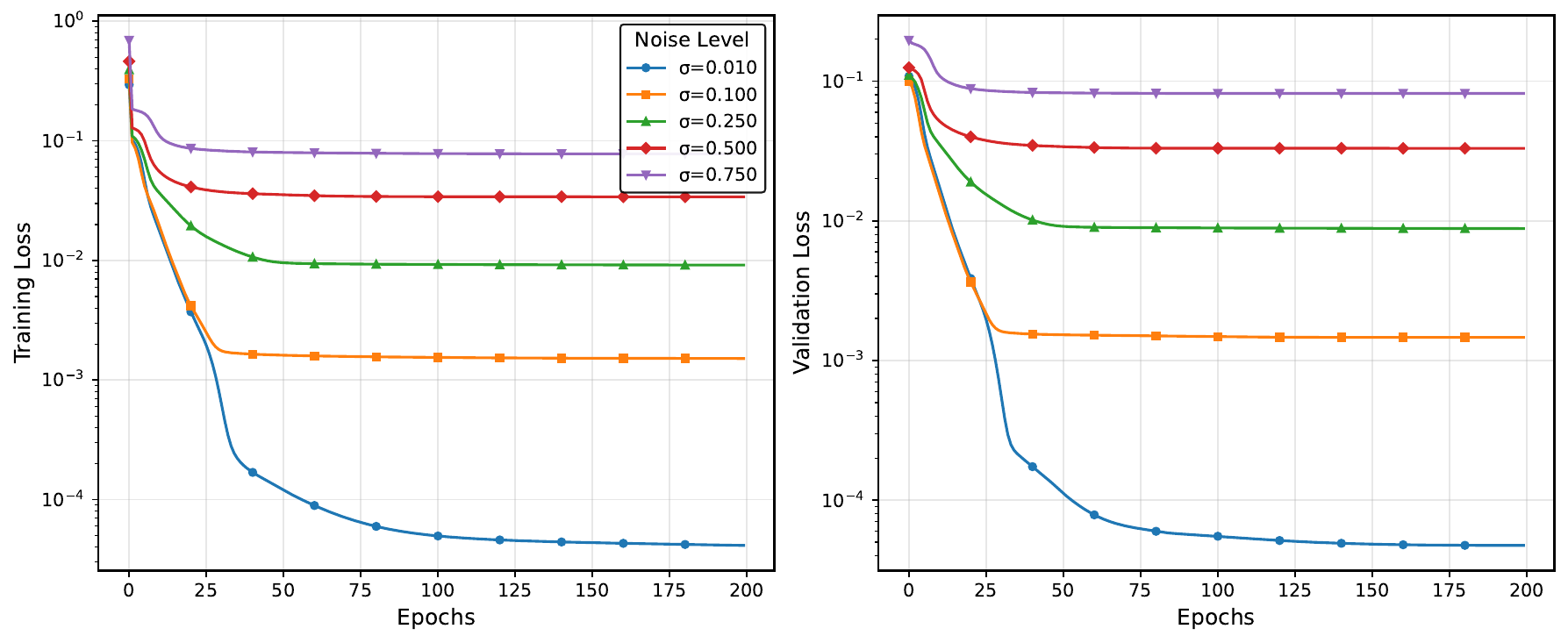}
    \caption{Epoch-wise training and validation loss convergence with different noise levels for the Burgers' equation}
    \label{fig:burgers_convergence}
\end{figure}

\begin{figure}[H]
    \centering
    \includegraphics[width=\linewidth]{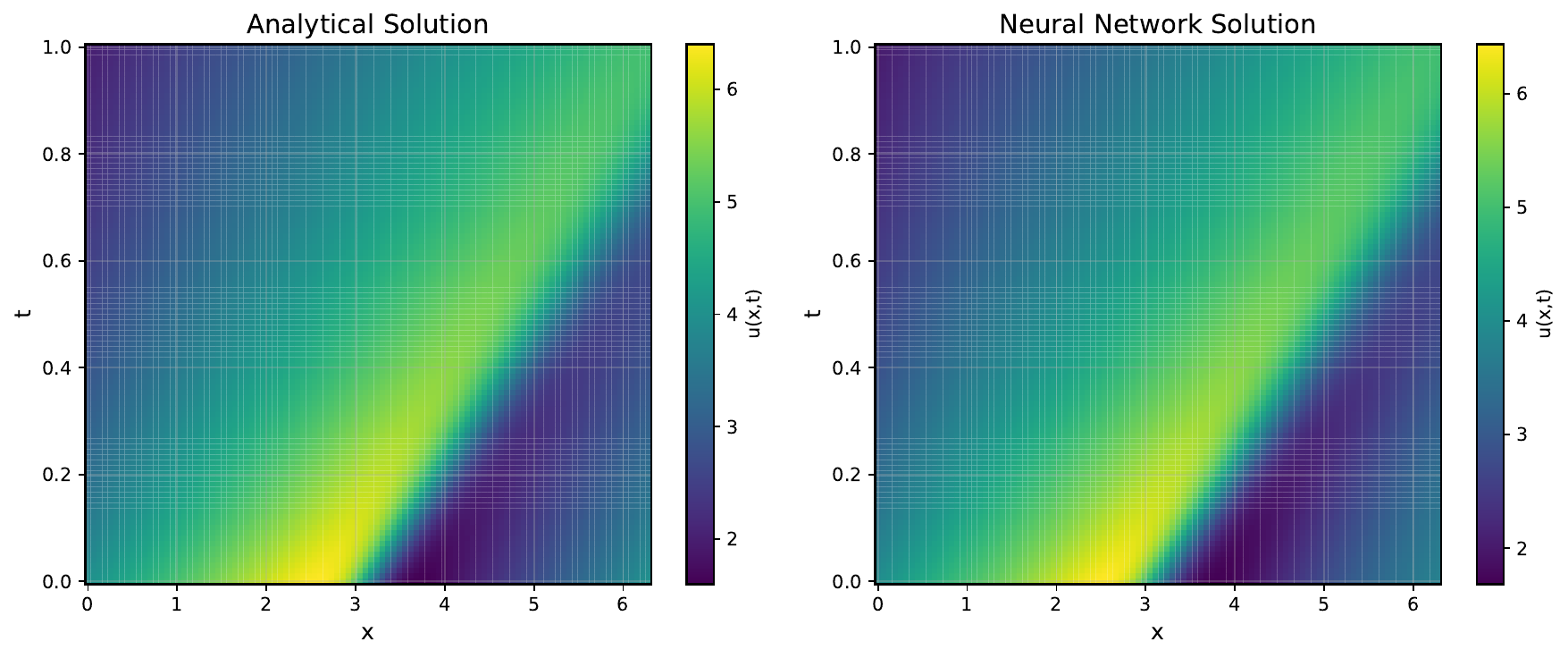}
    \caption{Real (left) vs. predicted (right) solution of the Burgers' equation. }
    \label{fig:burgers_nn_solution}
\end{figure}

In the training of Burgers' equation, the convergence of training and testing loss is shown in Fig.~\ref{fig:burgers_convergence}.
For all noise levels, the training and testing loss converge synchronously, indicating that the model is performing similarly well on both training and testing datasets.
We further visualize the predicted solution of the Burgers' equation in Fig.~\ref{fig:burgers_nn_solution}.
It is observable that even with limited scarce data, the shallow neural network can accurately predict the solution of the Burgers' equation.
Table~\ref{tab:burgers_error} and Table~\ref{tab:sindy_error_burgers} presents the error evolution of mesh-free SINDy from two different perspectives.
From Table~\ref{tab:burgers_error}, we discover that with larger sample sizes, the error of mesh-free SIDNy uniformly decreases and the model becomes more robust to noise.
In Table~\ref{tab:sindy_error_burgers}, we provide more evaluation metrics to assess the performance of mesh-free SINDy with increasing noise levels.

\begin{table}[H]
    \centering
    \caption{Loss of mesh-free SINDy under different sample sizes and noise levels for the Burgers' equation}
    \label{tab:burgers_error}
    \begin{tabular}{cccccc}
        \toprule
        \textbf{Samples} & \textbf{1\% Noise} & \textbf{10\% Noise} & \textbf{25\% Noise} & \textbf{50\% Noise} & \textbf{75\% Noise} \\
        \midrule
        100   & 0.6162 & 0.8487 & 1.2749 & 1.9100 & 2.2014 \\
        1000  & 0.2663 & 0.3753 & 0.5485 & 0.8724 & 1.0492 \\
        2500 & 0.1379 & 0.2822 & 0.4027 & 0.5942 & 0.7665 \\
        5000  & 0.1011 & 0.2214 & 0.3747 & 0.5030 & 0.6842 \\
        \bottomrule
    \end{tabular}
\end{table}

\begin{table}[H]
    \centering
    \caption{Error metrics of SINDy under different noise levels with $5000$ samples for the Burgers' equation}
    \label{tab:sindy_error_burgers}
    \begin{tabular}{lccccc}
        \toprule
        \textbf{Metrics} & \textbf{1\% Noise} & \textbf{10\% Noise} & \textbf{25\% Noise} & \textbf{50\% Noise} & \textbf{75\% Noise} \\
        \midrule
        $E_{\text{PDE}}$          & 4.76e-16 & 4.70e-16 & 4.88e-16 & 4.94e-16 & 4.75e-16 \\
        $E_{\text{NN}}$           & 0.0970   & 0.2912   & 0.5296   & 1.0727   & 1.4731   \\
        $E_{\text{dudt}}$         & 0.1095   & 0.2625   & 0.3242   & 0.4995   & 0.5658   \\
        $E_{\text{SINDy}}$        & 0.0966   & 0.2787   & 0.3001   & 0.5018   & 0.3995   \\
        $E_{\text{field}}$        & 0.1017   & 0.3318   & 0.5983   & 1.1713   & 1.4587   \\
        \bottomrule
    \end{tabular}
\end{table}

\subsection{Heat equation}
\label{app:heat}
\begin{figure}[H]
    \centering
    \includegraphics[width=\linewidth]{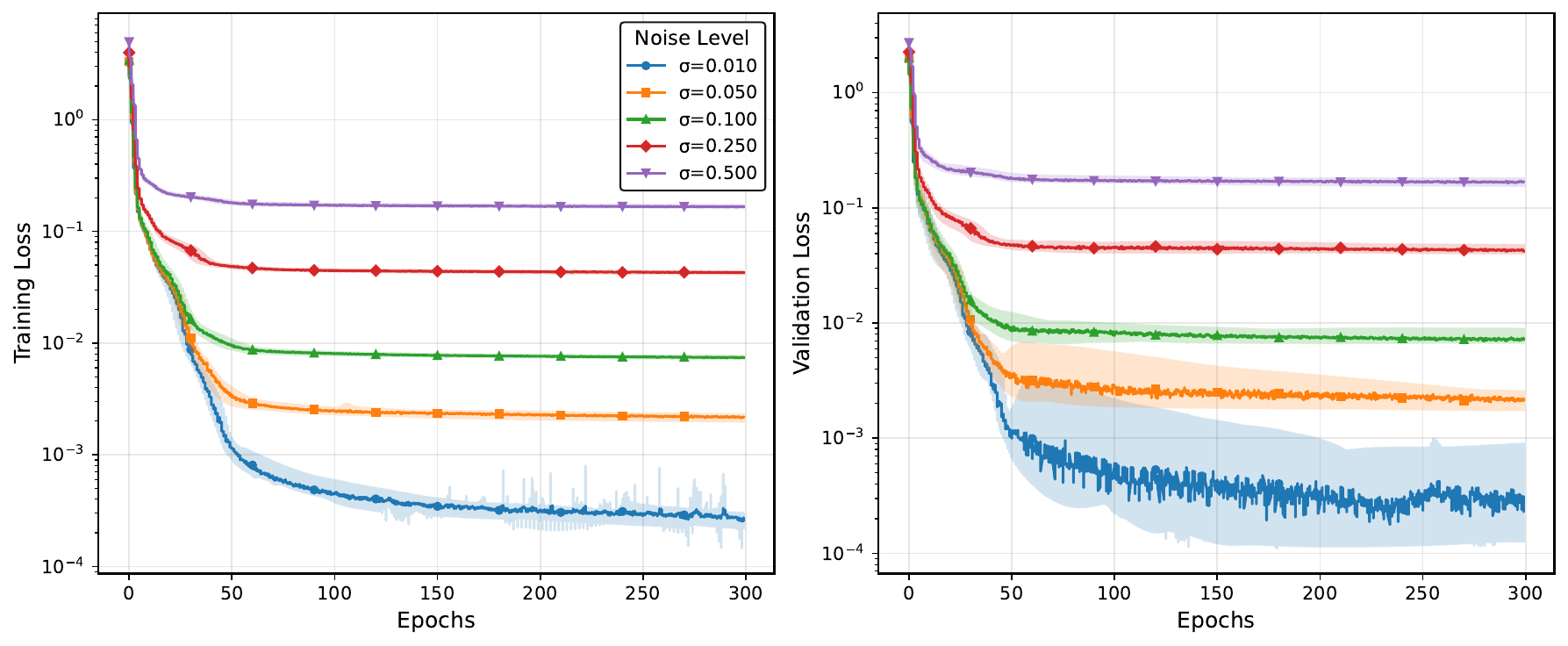}
    \caption{Epoch-wise training and validation loss convergence with different noise levels for the heat equation.}
    \label{fig:heat_convergence}
\end{figure}

\begin{figure}[H]
    \centering
    \includegraphics[width=\linewidth]{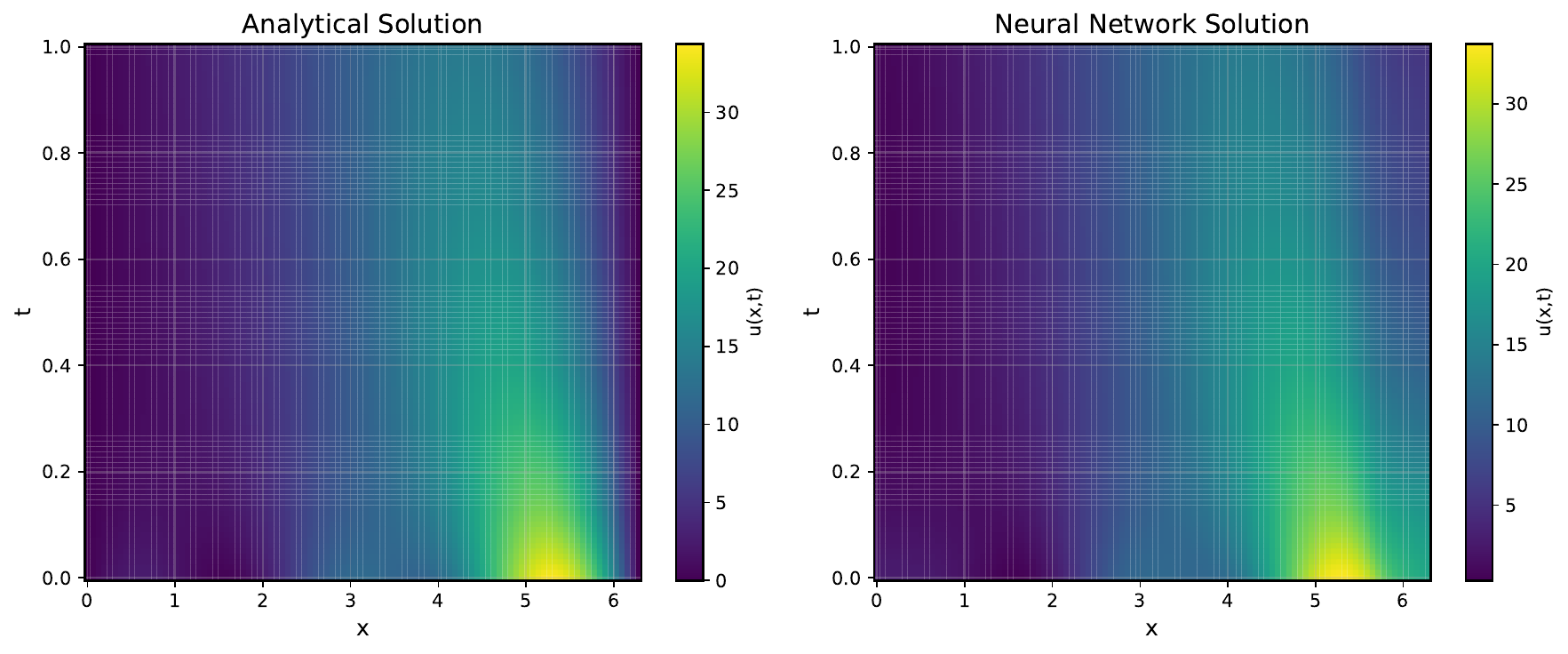}
    \caption{Real (left) vs. predicted (right) solution of the heat equation. }
    \label{fig:heat_nn_solution}
\end{figure}

In the training of the heat equation, the convergence of training and testing loss is shown in Fig.~\ref{fig:heat_convergence}.
For all noise levels, the training and testing loss converge synchronously, indicating that the model is performing similarly well on both training and testing datasets.
However, the testing loss demonstrates larger instability during training, which is likely due to the learning rate setting and choice of optimizer.
We further visualize the predicted solution of the heat equation in Fig.~\ref{fig:heat_nn_solution}.
The solution of heat equation is nicely predicted via neural network.
Table~\ref{tab:heat_error} and Table~\ref{tab:sindy_error_heat} presents the error evolution of mesh-free SINDy.
From Table~\ref{tab:heat_error}, we discover that with larger sample sizes, the error of mesh-free SIDNy uniformly decreases and the model becomes more robust to noise.
In Table~\ref{tab:sindy_error_heat}, we provide more evaluation metrics to assess the performance of mesh-free SINDy with increasing noise levels.

\begin{table}[H]
    \centering
    \caption{Loss of mesh-free SINDy under different sample sizes and noise levels for the heat equation.}
    \label{tab:heat_error}
    \begin{tabular}{ccccc}
        \toprule
        \textbf{Samples} & \textbf{1\% Noise} & \textbf{5\% Noise} & \textbf{10\% Noise} & \textbf{25\% Noise} \\
        \midrule
        500   & 5.1618 & 5.1329 & 5.2472 & 5.5408  \\
        1000  & 3.3018 & 3.6175 & 3.6939 & 5.2842 \\
        2500 & 2.6697 & 3.2724 & 4.7077 & 5.2627  \\
        4000  & 2.2628 & 3.7658 & 4.4749 & 5.0642  \\
        \bottomrule
    \end{tabular}
\end{table}

\begin{table}[H]
    \centering
    \caption{Error metrics of SINDy under different noise levels with $5000$ samples for the heat equation.}
    \label{tab:sindy_error_heat}
    \begin{tabular}{lccccc}
        \toprule
        \textbf{Metrics} & \textbf{1\% Noise} & \textbf{5\% Noise} & \textbf{10\% Noise} & \textbf{25\% Noise} & \textbf{50\% Noise} \\
        \midrule
        $E_{\text{PDE}}$          & 5.53e-07 & 5.56e-07 & 5.64e-07 & 5.63e-07 & 5.56e-07 \\
        $E_{\text{NN}}$           & 1.9749   & 4.3219   & 4.8943   & 11.7437  & 12.9245  \\
        $E_{\text{dudt}}$         & 0.4288   & 1.0015   & 0.9983   & 1.9481   & 2.9653   \\
        $E_{\text{SINDy}}$        & 2.3853   & 3.8856   & 5.1131   & 5.1594   & 5.3305   \\
        $E_{\text{field}}$        & 5.17e+04 & 6.09e+04 & 6.03e+04 & 8.99e+04 & 9.27e+04 \\
        \bottomrule
    \end{tabular}
\end{table}

\subsection{KdV equation}
\label{app:kdv}

\begin{figure}[H]
    \centering
    \includegraphics[width=\linewidth]{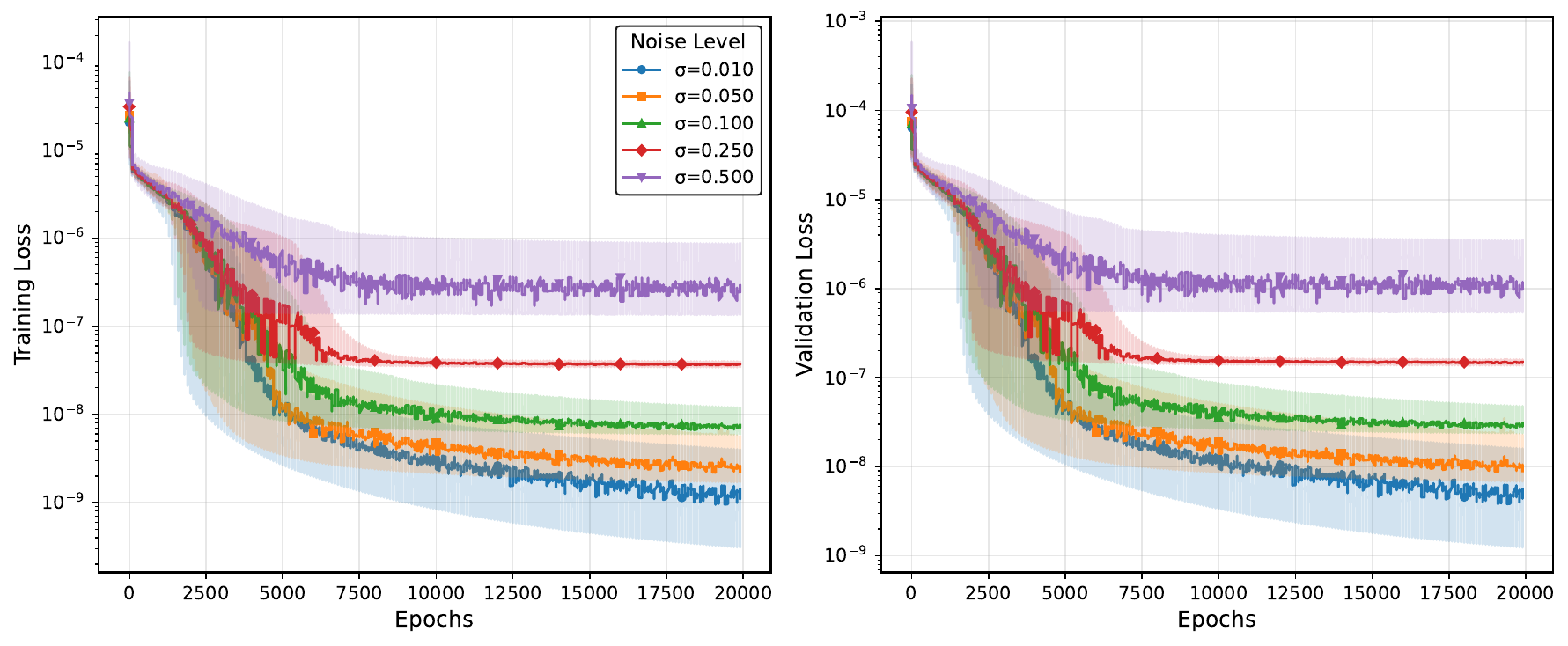}
    \caption{Epoch-wise training and validation loss convergence with different noise levels for the KdV equation.}
    \label{fig:kdv_convergence}
\end{figure}

\begin{figure}[H]
    \centering
    \includegraphics[width=\linewidth]{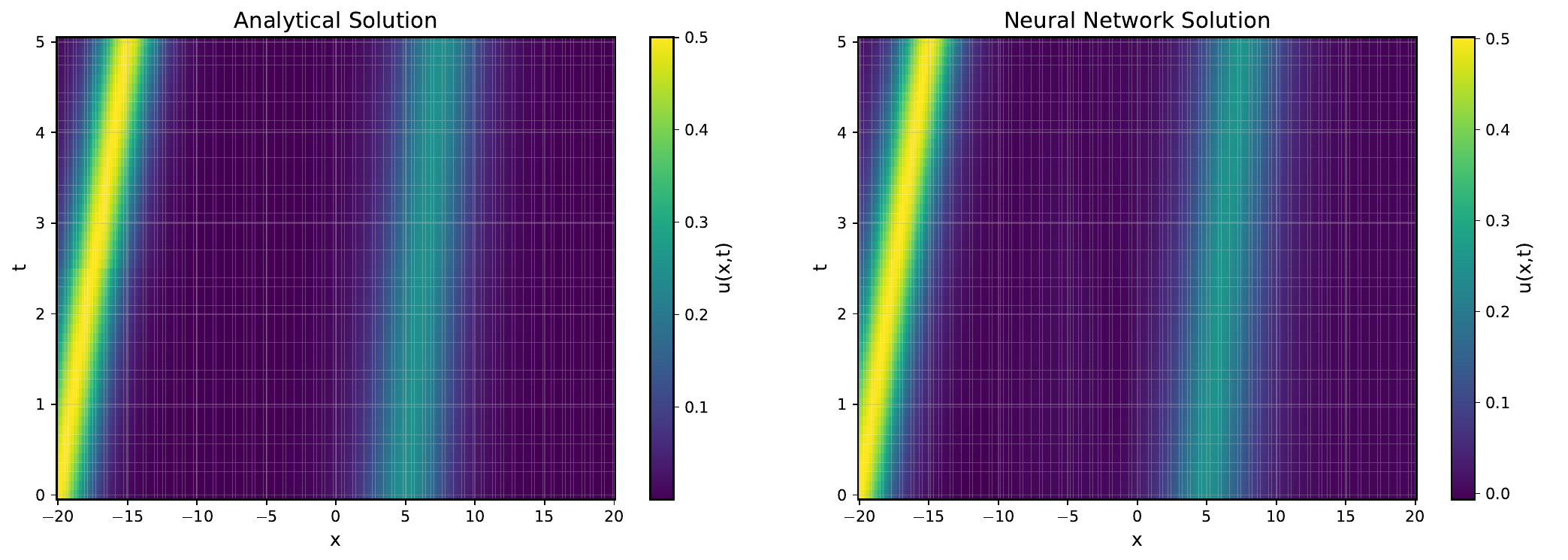}
    \caption{Real (left) vs. predicted (right) solution of the KdV equation}
    \label{fig:kdv_nn_solution}
\end{figure}

In the training of the KdV equation, the convergence of training and testing loss is shown in Fig.~\ref{fig:kdv_convergence}.
For all noise levels, the training and testing loss converge synchronously, indicating that the model is performing similarly well on both training and testing datasets.
Unlike the previous two equations, both training and testing loss demonstrates larger instability during training, which is likely due to the learning rate setting and choice of optimizer.
We further visualize the predicted solution of the KdV equation in Fig.~\ref{fig:kdv_nn_solution}.
The solution of KdV equation is nicely predicted via neural network.
Table~\ref{tab:kdv_error} and Table~\ref{tab:sindy_error_kdv} presents the error evolution of mesh-free SINDy.
From Table~\ref{tab:kdv_error}, we discover that with larger sample sizes, the error of mesh-free SIDNy uniformly decreases and the model becomes more robust to noise.
In Table~\ref{tab:sindy_error_kdv}, we provide more evaluation metrics to assess the performance of mesh-free SINDy with increasing noise levels.

\begin{table}[H]
    \centering
    \caption{Loss of mesh-free SINDy under different sample sizes and noise levels for the KdV equation.}
    \label{tab:kdv_error}
    \begin{tabular}{ccccc}
        \toprule
        \textbf{Samples} & \textbf{1\% Noise} & \textbf{5\% Noise} & \textbf{10\% Noise} & \textbf{25\% Noise} \\
        \midrule
        100   & 83.69 & 96.37 & 119.89 & 121.28  \\
        500   & 59.00 & 59.67 & 59.00 & 59.54  \\
        1000  & 60.09 & 59.30 & 59.22 & 59.62 \\
        2000 & 61.13 & 59.60 & 58.92 & 58.49  \\
        \bottomrule
    \end{tabular}
\end{table}

\begin{table}[H]
    \centering
    \caption{Error metrics equation under different noise levels with $1000$ samples for the KdV equation.}
    \label{tab:sindy_error_kdv}
    \begin{tabular}{lccccc}
        \toprule
        \textbf{Metrics} & \textbf{1\% Noise} & \textbf{5\% Noise} & \textbf{10\% Noise} & \textbf{25\% Noise} & \textbf{50\% Noise} \\
        \midrule
        $E_{\text{PDE}}$          & 1.76e-03 & 1.72e-03 & 1.68e-03 & 1.77e-03 & 1.65e-03 \\
        $E_{\text{NN}}$           & 41.9907  & 44.2752  & 42.6268  & 43.8201  & 44.7633  \\
        $E_{\text{dudt}}$         & 55.6690  & 56.4196  & 57.4353  & 58.6451  & 63.4641  \\
        $E_{\text{SINDy}}$        & 58.2436  & 56.9515  & 57.6004  & 59.1942  & 63.6559  \\
        $E_{\text{field}}$        & 43.1152  & 45.3267  & 43.6592  & 44.6639  & 45.4288  \\
        \bottomrule
    \end{tabular}
\end{table}

\subsection{Advection-diffusion equation}
\label{app:ad}
\begin{figure}[H]
    \centering
    \includegraphics[width=\linewidth]{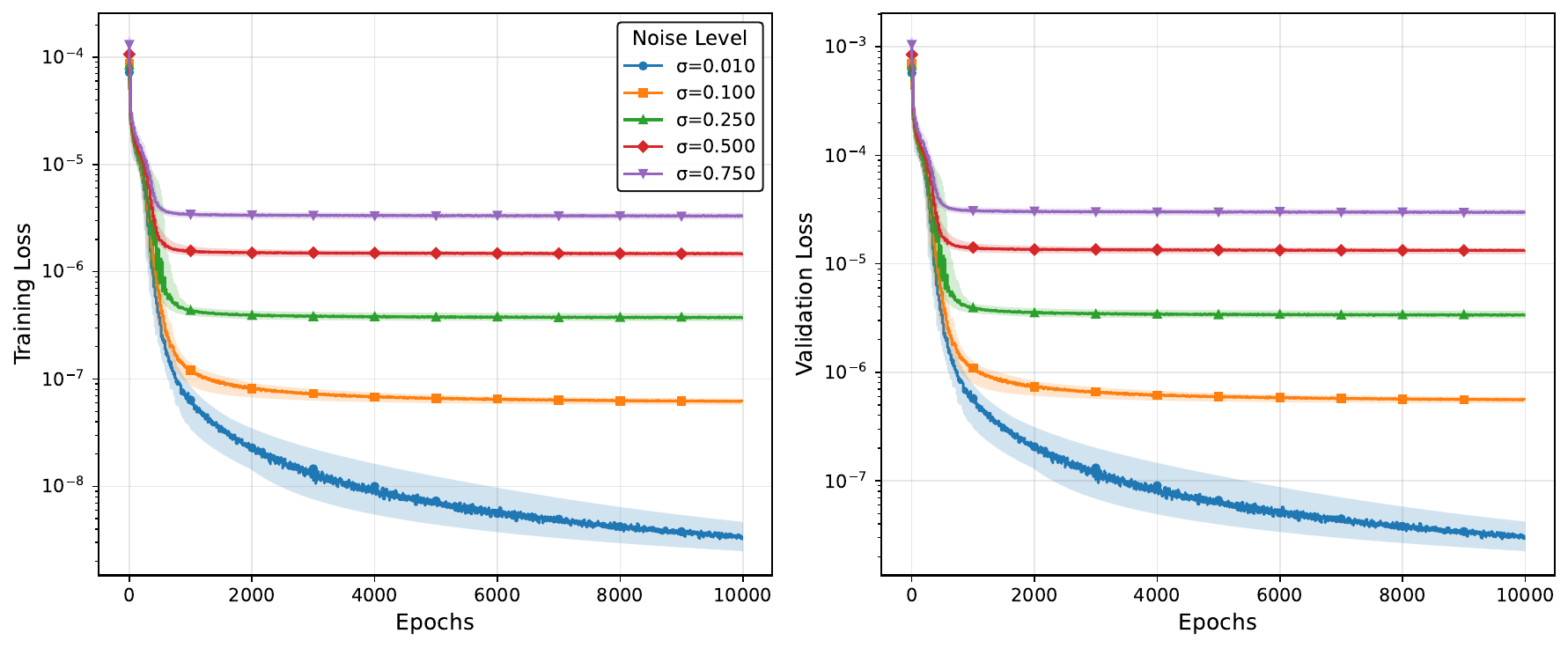}
    \caption{Epoch-wise training and validation loss convergence with different noise levels for the advection-diffusion equation.}
    \label{fig:ad_convergence}
\end{figure}

\begin{figure}[H]
    \centering
    \includegraphics[width=\linewidth]{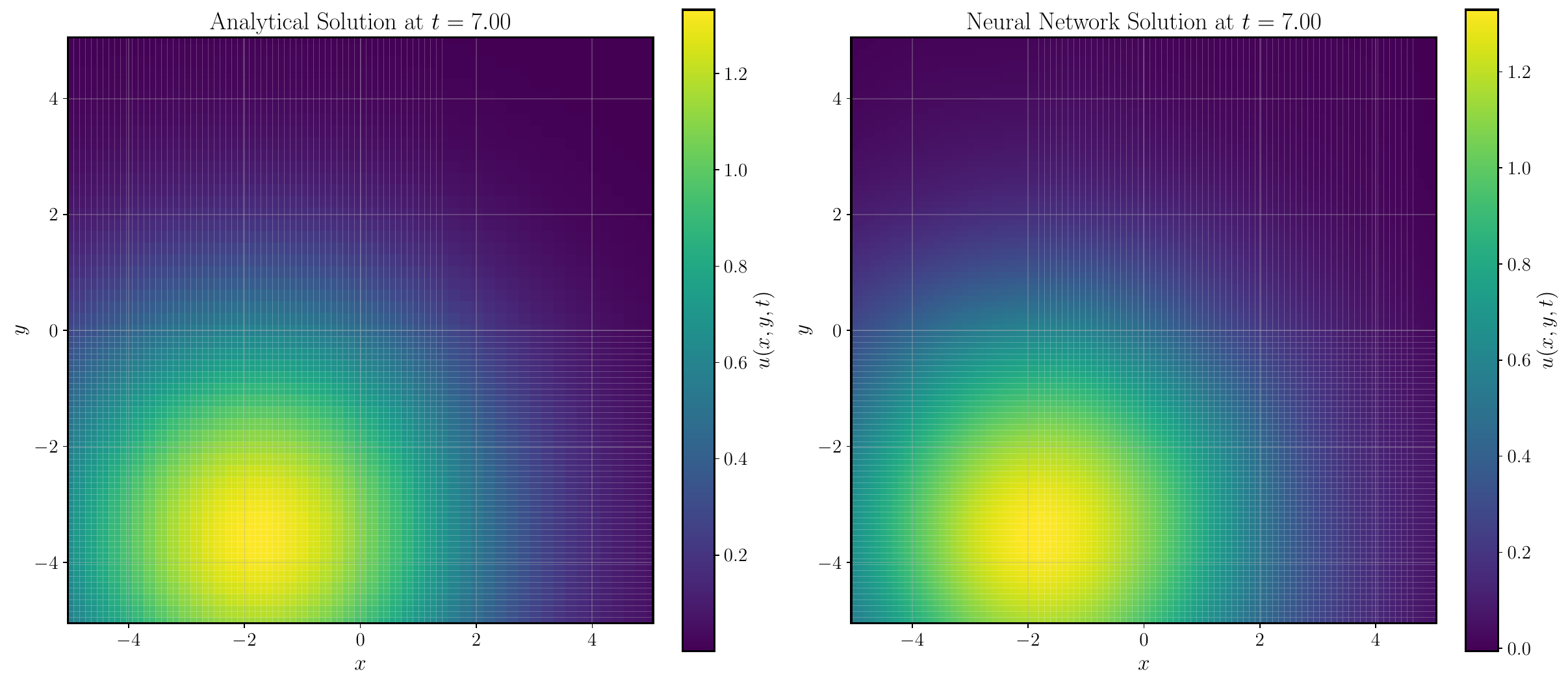}
    \caption{Real (left) vs. predicted (right) solution of the advection-diffusion equation. }
    \label{fig:ad_nn_solution}
\end{figure}

In the training of the advection-diffusion equation, the convergence of training and testing loss is shown in Fig.~\ref{fig:ad_convergence}.
For all noise levels, the training and testing loss converge synchronously, indicating that the model is performing similarly well on both training and testing datasets.
Since we uniformly set the learning rate for all experiments, the training and testing loss for $1\%$ noise level demonstrates minor instability.
We further visualize the predicted solution of the advection-diffusion equation in Fig.~\ref{fig:ad_nn_solution}.
The solution of advection-diffusion equation is nicely predicted via neural network.
Table~\ref{tab:ad_error} and Table~\ref{tab:sindy_error_ad} presents the error evolution of mesh-free SINDy.
From Table~\ref{tab:ad_error}, we discover that with larger sample sizes, the error of mesh-free SIDNy uniformly decreases and the model becomes more robust to noise.
In Table~\ref{tab:sindy_error_ad}, we provide more evaluation metrics to assess the performance of mesh-free SINDy with increasing noise levels.

\begin{table}[H]
    \centering
    \caption{Loss of mesh-free SINDy under different sample sizes and noise levels for the advection-diffusion equation.}
    \label{tab:ad_error}
    \begin{tabular}{ccccc}
        \toprule
        \textbf{Samples} & \textbf{1\% Noise} & \textbf{10\% Noise} & \textbf{25\% Noise} & \textbf{50\% Noise} \\
        \midrule
        500  & 905.87 & 919.62 & 920.59 & 1003.33 \\
        2000 & 916.12 & 916.99 & 923.89 & 922.10  \\
        4000 & 913.72 & 924.95 & 951.24 & 951.34  \\
        6000 & 929.11 & 951.23 & 906.92 & 934.35  \\
        \bottomrule
    \end{tabular}
\end{table}

\begin{table}[H]
    \centering
    \caption{Error metrics of SINDy under different noise levels with $6000$ samples for the advection-diffusion equation.}
    \label{tab:sindy_error_ad}
    \begin{tabular}{lccccc}
        \toprule
        \textbf{Metrics} & \textbf{1\% Noise} & \textbf{10\% Noise} & \textbf{25\% Noise} & \textbf{50\% Noise} & \textbf{75\% Noise} \\
        \midrule
        $E_{\text{PDE}}$          & 1.42e+03 & 1.45e+03 & 1.38e+03 & 1.40e+03 & 1.39e+03 \\
        $E_{\text{NN}}$            & 0.1896   & 0.1915   & 0.1834   & 0.1889   & 0.1905   \\
        $E_{\text{dudt}}$          & 931.10   & 951.35   & 904.72   & 936.53   & 926.19   \\
        $E_{\text{SINDy}}$        & 929.11   & 951.23   & 906.92   & 934.35   & 917.25   \\
        $E_{\text{field}}$             & 4.03e+03 & 4.08e+03 & 4.03e+03 & 4.26e+03 & 4.40e+03 \\
        \bottomrule
    \end{tabular}
\end{table}

\end{document}